\documentclass{Interspeech}
\usepackage{multirow}

\interspeechcameraready

\usepackage{tikz}
\usepackage{eso-pic}

\AddToShipoutPictureBG*{
  \AtPageUpperLeft{
    \raisebox{-2.5cm}{\makebox[\paperwidth]{\hfill\textit{\large{\textcolor{blue}{Accepted to Interspeech 2025}}}\hfill}}
  }
}

\title{Multi-Teacher Language-Aware Knowledge Distillation for Multilingual Speech Emotion Recognition}

\author[affiliation={}]{Mehedi Hasan}{Bijoy}
\author[affiliation={}]{Dejan}{Porjazovski}
\author[affiliation={}]{Tamás}{Grósz}
\author[affiliation={}]{Mikko}{Kurimo}

\affiliation[nocounter]{Department of lnformation and Communications Engineering}{Aalto University}{Finland}

\email{firstname.lastname@aalto.fi}


\usepackage{comment}

\begin{document}

\maketitle

\begin{abstract}
    
    

    Speech Emotion Recognition (SER) is crucial for improving human-computer interaction.
    Despite strides in monolingual SER, extending them to build a multilingual system remains challenging. Our goal is to train a single model capable of multilingual SER by distilling knowledge from multiple teacher models.
    To address this, we introduce a novel language-aware multi-teacher knowledge distillation method to advance SER in English, Finnish, and French. It leverages Wav2Vec2.0 as the foundation of monolingual teacher models and then distills their knowledge into a single multilingual student model. The student model demonstrates state-of-the-art performance, with a weighted recall of 72.9 on the English dataset and an unweighted recall of 63.4 on the Finnish dataset, surpassing fine-tuning and knowledge distillation baselines. Our method excels in improving recall for sad and neutral emotions, although it still faces challenges in recognizing anger and happiness.
\end{abstract}

\section{Introduction}
Speech Emotion Recognition (SER) involves identifying and analyzing emotional states in spoken languages. In multilingual settings, it becomes challenging due to the distinct linguistic features of each language. Unlike simple fine-tuning (FT) and knowledge distillation (KD), multi-teacher knowledge distillation (MTKD) specifically addresses this challenge by using multiple teacher models, each specialized in a different language, to enhance the student model's performance. The importance of SER is recognized as it benefits applications in human-computer interaction and speech processing tasks by fostering empathetic and effective interactions. For instance, SER aids in early detection and intervention in mental health issues by monitoring speech~\cite{elsayed2022speech}, even in multilingual communities.

Recent efforts in SER have primarily focused on leveraging pre-trained transformer-based large audio-language models due to their robust speech representation capabilities \cite{grosz2023discovering}. Moreover, it is hypothesized that integrating knowledge from multiple teacher models significantly enhances student model performance by utilizing complementary insights. For example, Confidence-Aware MTKD dynamically adapts to each teacher's reliability \cite{zhang2022confidence}, while Adaptive MTKD employs a meta-weight network to coordinate diverse knowledge from multiple teachers \cite{zhang2023adaptive}. Their empirical results indicate that these methods outperform FT and KD models, particularly in low-resource settings \cite{anand2023multi}. Overall, there is a clear shift towards adaptive meta-learning strategies, pre-trained models, and MTKD to achieve high performance and efficiency.

We found that previous studies have not investigated MTKD for multilingual SER, 
particularly in training a multilingual student model from monolingual teachers.
Therefore, we hypothesize that optimizing MTKD for multilingual SER can enhance emotional knowledge transfer across languages. 
By leveraging cross-lingual data,
we aim to better integrate emotional knowledge in the student model. 
Our objective is to construct a single multilingual SER model through the distillation of knowledge obtained from multiple monolingual teacher models.
We investigate whether MTKD improves cross-linguistic emotional knowledge transfer in multilingual SER.
To achieve this, we use cosine similarity scores to select the most suitable teacher for each language.
Consequently, this targeted approach should improve cross-linguistic knowledge integration, allowing better generalization across languages. 
By mitigating score variation through cosine similarity rescaling, our method enhances stability, resulting in improved cross-linguistic generalization and reliable performance, even in challenging multilingual scenarios.


The main contributions of this study are summarized below:
\begin{itemize}
    \item We propose a language-aware MTKD method using three teacher models to enhance SER in English, Finnish, and French
    It includes both monolingual and multilingual configurations, setting a novel benchmark for multilingual SER.
    \item Our proposed method achieves superior performance across datasets by surpassing different training paradigms, such as standard FT and conventional KD baselines, demonstrating generalizability and robustness.
\end{itemize}

\section{Related Works}\label{sec:Literature_Review}


Numerous approaches have been developed for the SER task, which can broadly be categorized into four groups: simple FT, conventional KD, MTKD, and multimodal fusion. 
As for simple FT, pre-trained models like wav2vec2.0 \cite{baevski2020wav2vec}, HuBERT \cite{hsu2021hubert}, and WavLM \cite{chen2022wavlm} have significantly advanced SER by addressing the issue of data scarcity \cite{pepino2021emotion}. Leveraging transfer learning and FT, as demonstrated in the works of \cite{grosz2022wav2vec2} and \cite{chen2023exploring}, markedly improved the performance of emotion recognition from speech. Sharma et al. \cite{sharma2022multi} expanded on this by developing a multilingual learning system, while another study introduced the P-TAPT method to better align pre-training with target tasks \cite{chen2023exploring}. Unlike the previous studies, Chakhtouna et al. \cite{chakhtouna2024unveiling} proposed a model combining HuBERTX-large with an SVM to emphasize the importance of advanced feature extraction. However, while the effectiveness of fine-tuning is supported by the empirical outcomes of \cite{grosz2022wav2vec2} and \cite{chen2023exploring}, limitations such as potential overfitting and model convergence issues are observed in \cite{chakhtouna2024unveiling}.

When it comes to conventional KD, studies have expanded its scope for SER by transferring knowledge from a larger teacher model to a smaller student model to optimize performance \cite{gao2022multi}. For example,  Hao et al. \cite{hao2024one} introduced the OFA-KD framework for heterogeneous architectures, and Zhao et al. \cite{zhao2023hierarchical} proposed a hierarchical network with decoupled KD. Similarly, another work focused on KD-based model adaptation for emotional speech \cite{yun2023end}. 
These studies generally aim to enhance the model performance. 
However, challenges such as the effectiveness of dropout \cite{sridhar2020ensemble}, resource demands \cite{takashima2018investigation}, and the need for improvements in multi-modal KD \cite{zhao2023hierarchical} are still persistent.

With respect to MTKD, several methods have been proposed, including a confidence-aware MTKD framework to address low-performing teachers \cite{zhang2022confidence}, switched-training for different resource settings \cite{fukuda2017efficient}, and adaptive MTKD with meta-learning \cite{zhang2023adaptive}. In contrast, Yang et al. \cite{yang2020model} and Ren et al. \cite{ren2023fast} concentrated on efficiency and resource optimization through two-stage MTKD and self-distillation, respectively. Additionally, a few studies explored multimodal and ensemble methods for enhancing performance \cite{anand2023multi, huang2023ensemble}. Despite these advancements, challenges such as high computational demands \cite{zhang2023adaptive, anand2023multi}, large dataset requirements \cite{yang2020model}, and increased complexity \cite{fukuda2017efficient, ren2023fast} persist, highlighting ongoing limitations in the field.

\section{Methodology}\label{sec:Methodology}



Our proposed method involves training a student model $S$ guided by three teacher models $T_1$, $T_2$, and $T_3$. The training process uses the raw audio waveform input $\mathbf{X}$, which is a discrete-time sequence of $N$ successive samples $\{x[1], x[2], \dots, x[N]\}$ such that $x_t \in \mathbb{R}$ . Three teacher models, $T_1$, $T_2$, and $T_3$, and the student model $S$ each process $\mathbf{X}$ and produce logits for $K$ classes. The Kullback-Leibler (KL) divergence between the student logits and each teacher's logits is computed as $\mathcal{L}_{KL}^{i}(\sigma(l_{T_i}) \parallel \sigma(l_{S}))$, where $\sigma$ is the softmax function. The total KL divergence loss is $\mathcal{L}_{KL} = \sum_{i=1}^{3} c_i \times \mathcal{L}_{KL}^{i}$, with $c_i$ being constants derived from cosine similarities. The cross-entropy (CE) loss between the student logits and true labels $\mathbf{y}$ is $\mathcal{L}_{CE} (\sigma(l_{S}), \mathbf{y})$. The total loss is calculated as: 
$\mathcal{L} = (1-\lambda)\mathcal{L}_{CE} + \lambda \mathcal{L}_{KL}$, where $\lambda$ balances the contributions of CE and KL divergence losses, enabling the model to learn accurate classification while aligning its predicted distribution with the targeted teacher's distribution.



\begin{figure}[t!]
  \begin{center}
    \includegraphics[width=\columnwidth]{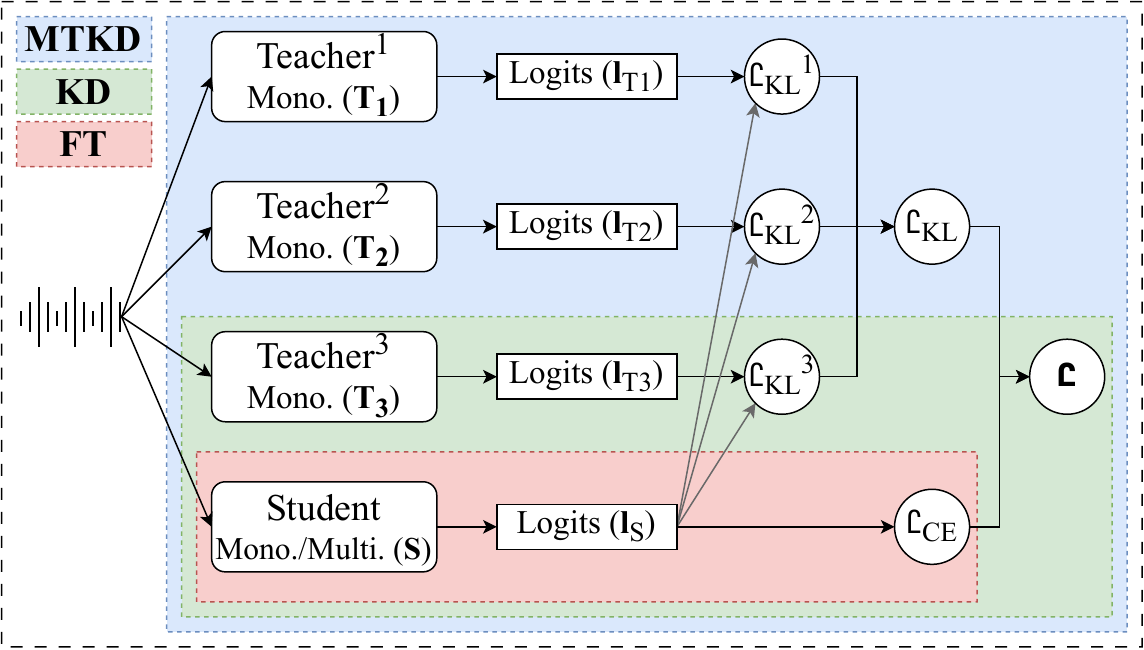} 
  \end{center}
  \caption{
  Proposed language-aware MTKD method.
  } 
  \label{fig:mtkd_method}
\end{figure}

\noindent\textbf{Motivations:}
Our proposed method advances SER by addressing cross-language variability, a key challenge in multilingual SER tasks that hinders generalization. We introduce a novel language-aware MTKD method, where multiple teacher models, each specialized in a different language, distill language-specific emotional cues into a student model. Using cosine similarity scores, the student model captures nuanced emotional patterns for each language while learning cross-linguistic representations. This approach improves generalizability and multilingual performance, setting a new benchmark compared to prior methods that lack language-specific optimization. Figure~\ref{fig:mtkd_method} presents an overview of our proposed MTKD method.

\subsection{Proposed Language-Aware MTKD Method}

The proposed method utilizes three pre-trained teacher models, $\mathbf{T_i(.)}$, where $i \in \{1, 2, 3\}$, each specializing in English, Finnish, and French, respectively, alongside a student model $\mathbf{S(.)}$ that can be monolingual or multilingual. The raw audio waveform $\mathbf{X}$ is fed into each model, producing logits $l_{T_1}$, $l_{T_2}$, $l_{T_3}$, and $l_{S}$, where each set of logits corresponds to $K$ number of classes. Then, cosine similarities between the student's and each teacher's logits are calculated to measure alignment:
{\footnotesize
\begin{equation}
    \text{cs}_i = \frac{\sum_{j=1}^{K} \mathbf{l}_{s}[j] \mathbf{l}_{T_i}[j]}{\sqrt{\sum_{j=1}^{K} (\mathbf{l}_{s}[j])^2} \sqrt{\sum_{j=1}^{K} (\mathbf{l}_{T_i}[j])^2}} \quad \text{for } \; i \in \{1, 2, 3\}
    \label{cosine_sim}
\end{equation}
}

To prioritize the most relevant teacher, these cosine similarity scores are scaled by a small temperature value $\mathbf{\tau}<1.0$ and transformed via a softmax function, enhancing small differences and focusing on the most relevant teacher: 
$\text{cs}'_i = {\exp\left({\text{cs}_i}/{\tau}\right)} / {\sum_{k=1}^{3} \exp\left({\text{cs}_k}/{\tau}\right)}$. 
Scaling the cosine similarity by dividing by a small $\tau$ sharpens the differences between similarity scores, as $\exp(\text{cs}_i/\tau)$ amplifies higher scores more strongly, making the most relevant teacher's contribution more dominant.
The student and teacher logits are then converted into probabilities using the softmax function with a higher temperature $\mathbf{\tau}>1.0$, such that: 
{
\footnotesize
\begin{equation}
    P_{x}[j] = \frac{\exp\left(\frac{\mathbf{l}{x}[j]}{T}\right)}{\sum_{k=1}^{K} \exp\left(\frac{\mathbf{l}_{x}[k]}{T}\right)} ; for \ x \in \{S, T_i \}
\end{equation}
}
Dividing the logits by a high temperature value $\tau$ smooths the probability distribution, lowering the model's confidence in specific classes and fostering exploration by reducing the differences between logits of those $K$ classes.
%
%
Next, the KL divergence loss is computed to ensure the student model mimics the teachers' probability distributions: 
%
{
\footnotesize
\begin{equation}
    \mathcal{L}_{KL}^{i} = \sum_{j=1}^{K} P_{S}[j] \log \left(\frac{P_{S}[j]}{P_{T_i}[j]}\right) \quad \text{for } \; i \in \{1, 2, 3\}
    \label{kl_teachers}
\end{equation}
}

%
%
Moreover, we aim to ensure that the student model prioritizes learning from the most relevant teacher. To do so, we combine the distribution errors from the three different teachers by weighting them with the enhanced cosine similarity scores of each teacher such that $\mathcal{L}_{KL} = \sum_{i=1}^{3} \text{cs}'_i \times \mathcal{L}_{KL}^{i}$.
Next, the CE loss is calculated to measure how well the student predicts the classes by comparing the student's predicted probabilities with corresponding class labels (eq. \ref{ce}).
{
\footnotesize
\begin{equation}
    \mathcal{L}_{CE} = -\sum_{j=1}^{K} y_j \log\left(\frac{\exp(\mathbf{l}_S[j])}{\sum_{k=1}^{K} \exp(\mathbf{l}_S[k])}\right)
    \label{ce}
\end{equation}
}
Finally, to enable language-aware distillation, we combine the classification error $\mathcal{L}_{CE}$ with the distribution error $\mathcal{L}_{KL}$ to calculate $\mathcal{L}$. This ensures that the model's prediction logit distribution is similar to the appropriate teacher and that it correctly classifies the actual labels,
ensuring that the student model not only aligns its output distribution with the most relevant teacher, but also accurately classifies the actual labels.

\section{Experimental Setup}
\noindent\textbf{Evaluation Metrics:} To accommodate the varying evaluation metrics used in different studies, we report UR \cite{ren2023fast}, WR \cite{sharma2022multi}, unweighted accuracy (UA) \cite{chen2023exploring}, and weighted accuracy (WA)  \cite{chen2024vesper} to ensure comprehensive results and facilitate cross-study comparisons. 
UR and UA ignore the class distribution, while WR and WA adjust for it.

\noindent\textbf{Datasets:}
We utilize three datasets: IEMOCAP \cite{busso2008iemocap} for English, FESC \cite{airas2006emotions} for Finnish, and CaFE \cite{gournay2018canadian} for French. We focus on four common emotion classes, including angry, happy, neutral, and sad, which are available in all three of these datasets. This eliminates class inconsistency issue in the multilingual setup. The statistics for the considered portion of these datasets can be found in Table \ref{datasets}.









\noindent\textbf{Baselines:}
FT-Mono. fine-tunes Wav2Vec2.0-base\footnote{https://huggingface.co/facebook/wav2vec2-base} on a monolingual dataset, while FT-Multi. fine-tunes same Wav2Vec2.0-base$^1$ model on a multilingual dataset. Additionally, KD-Mono. is a knowledge distillation approach where a monolingual student model (Wav2Vec2.0-base$^1$) learns from a multilingual teacher model, which is fine-tuned FT-Multi.

\noindent\textbf{Cross Validation:}
In the multilingual SER task, we use predefined splits from the English and Finnish datasets and a single split from the French dataset to train and evaluate our language-aware MTKD model on non-overlapping sets.

\noindent\textbf{Experiments:}
For our experiments, we use three teacher models specialized in English, Finnish, and French SER tasks, which are fine-tuned \textit{Wav2Vec2-base}$^{1}$ models, trained for 20 epochs with a learning rate of $3e^{-5}$ and a batch size of 32.
In our language-aware MTKD method, processed audio is used as an input to each model to produce soft outputs. 
We distill knowledge from the teacher models to the student model by calculating the KL divergence loss, comparing the student's predictions with each teacher's predictions.
The losses are combined using amplified cosine similarity scores and a softmax function to prioritize the correct language. Outputs are further smoothed using a temperature value of 5. The final student model is optimized by combining cross-entropy loss and KL divergence loss, weighted at 75\% and 25\% respectively, ensuring effective learning of both general and language-specific SER tasks. The code for reproducing both the proposed method and the baselines is publicly available at \href{https://github.com/aalto-speech/mtkd4ser}{\texttt{https://github.com/aalto-speech/mtkd4ser}}.

\begin{table}[t!]
\centering
\caption{The composition of combined multilingual data.}
\resizebox{\columnwidth}{!}{
\begin{tabular}{ccccc}
\hline
\multirow{2}{*}{Dataset} & \multirow{2}{*}{Language} & \multirow{2}{*}{Splits} & \multicolumn{1}{c}{Train} & \multicolumn{1}{c}{Test} \\ 
 &  &  & $\#$ Samples & $\#$ Samples  \\ \hline

{IEMOCAP} & {English} & {5} & {$\approx$4508} & {$\approx$1241} \\ 

{FESC} & {Finnish} & {9} & {$\approx$2798} & {$\approx$461} \\ 

{CaFE} & {French} & {1} & {420} & {84}\\ \hline 

\end{tabular}
}

\label{datasets}
\end{table}

\section{Results}

\subsection{Quantitative Analysis}

\subsubsection{Performance on IEMOCAP}
Table \ref{performance_on_iemocap} presents the empirical results of various training approaches, including our proposed MTKD method, on the IEMOCAP dataset. This comparison highlights the effectiveness and generalizability of three training paradigms: fine-tuning, knowledge distillation, and multi-teacher knowledge distillation.
\renewcommand{\arraystretch}{1.25}
\begin{table*}[ht]
\centering
\caption{Performance of our proposed MTKD methods alongside other baselines on the IEMOCAP dataset, where 'Mono.' refers to monolingual, and 'Multi.' stands for multilingual configurations. The upper (numerator) and lower (denominator) bounds of the confidence interval are reported for both UR and WR.}
\begin{tabular}{ccccccccccc}
\hline
\multirow{2}{*}{Split} & \multicolumn{2}{|c|}{FT-Mono.} & \multicolumn{2}{c|}{FT-Multi.} & 
\multicolumn{2}{c|}{KD-Mono.} & \multicolumn{2}{c|}{\textbf{MTKD-Mono.}} & \multicolumn{2}{c|}{\textbf{MTKD-Multi.}} \\ 
\cline{2-11}
{} & {UR} & {WR} & {UR} & {WR} & {UR} & {WR} & {UR} & {WR} & {UR} & {WR} \\ \hline

{Split 1} & {69.8 \( \frac{74.5}{64.4} \)} & {65.8 \( \frac{70.6}{65.8} \)} & {70.6 \( \frac{74.1}{66.7} \)} & {70.7 \( \frac{74.0}{67.2} \)} & {72.5 \( \frac{77.4}{66.9} \)} & {70.3 \( \frac{75.1}{64.9} \)} & {72.4 \( \frac{76.8}{67.1} \)} & {68.5 \( \frac{73.1}{63.2} \)} & {\textbf{72.5} \( \frac{76.0}{68.7} \)} & {\textbf{73.4} \( \frac{76.6}{69.9} \)} \\ 

{Split 2} & {\textbf{79.3} \( \frac{83.9}{73.3} \)} & {\textbf{78.1} \( \frac{82.6}{72.8} \)} & {70.9 \( \frac{74.4}{67.3} \)} & {74.9 \( \frac{77.9}{71.6} \)} & {76.7 \( \frac{81.5}{71.0} \)} & {74.8 \( \frac{79.5}{69.3} \)} & {77.8 \( \frac{82.4}{72.1} \)} & {76.0 \( \frac{80.5}{70.6} \)} & {71.2 \( \frac{74.8}{67.3} \)} & {71.0 \( \frac{74.4}{67.3} \)} \\ 

{Split 3} & {66.2 \( \frac{71.3}{60.7} \)} & {65.9 \( \frac{71.0}{60.4} \)} & {67.7 \( \frac{71.2}{64.0} \)} & {69.5 \( \frac{72.8}{66.0} \)} & {69.5 \( \frac{74.5}{64.0} \)} & {69.1 \( \frac{74.1}{63.6} \)} & {68.9 \( \frac{73.9}{63.5} \)} & {69.0 \( \frac{74.0}{63.6} \)} & {\textbf{71.4} \( \frac{74.9}{67.8} \)} & {\textbf{71.5} \( \frac{74.8}{67.9} \)} \\ 

{Split 4} & {68.5 \( \frac{73.9}{62.7} \)} & {70.2 \( \frac{75.1}{64.8} \)} & {64.2 \( \frac{68.0}{60.3} \)} & {69.7 \( \frac{73.0}{66.1} \)} & {70.2 \( \frac{75.3}{64.3} \)} & {70.1 \( \frac{75.0}{64.7} \)} & {\textbf{72.3} \( \frac{77.6}{66.3} \)} & {72.8 \( \frac{77.8}{67.2} \)} & {71.6 \( \frac{75.3}{67.7} \)} & {\textbf{73.8} \( \frac{77.0}{70.3} \)} \\ 

{Split 5} & {72.2 \( \frac{76.9}{66.8} \)} & {70.5 \( \frac{75.1}{65.4} \)} & {67.9 \( \frac{71.3}{64.2} \)} & {70.2 \( \frac{73.4}{66.8} \)} & {72.2 \( \frac{77.0}{66.7} \)} & {71.2 \( \frac{75.8}{66.0} \)} & {71.0 \( \frac{75.9}{65.5} \)} & {69.7 \( \frac{74.4}{64.5} \)} & {\textbf{73.0} \( \frac{76.3}{69.3} \)} & {\textbf{74.7} \( \frac{77.8}{71.3} \)} \\  \hline

{Mean} & {71.2 \( \frac{76.1}{65.6} \)} & {70.1 \( \frac{74.9}{65.8} \)} & {68.2 \( \frac{71.8}{64.5} \)} & {71.0 \( \frac{74.2}{67.6} \)} & {{72.2} \( \frac{77.1}{66.6} \)} & {71.1 \( \frac{76.1}{65.8} \)} & {\textbf{72.5} \( \frac{77.3}{66.9} \)} & {71.2 \( \frac{76.0}{65.8} \)} & {{71.9} \( \frac{75.5}{68.2} \)} & {\textbf{72.9} \( \frac{76.1}{69.3} \)} \\  \hline

\end{tabular}

\label{performance_on_iemocap}
\end{table*}
The results demonstrate that the MTKD method with a monolingual student and monolingual teachers (\textit{MTKD-Mono.}) achieves the highest mean UR of 72.5, while the MTKD method with a multilingual student and monolingual teachers (\textit{MTKD-Multi.}) achieves the highest mean WR of 72.9. 
While other methods exhibit variability in performance and achieve superiority only in specific splits, our proposed \textit{MTKD-Multi.} method consistently outperforms them across most splits, underscoring its robustness.



\subsubsection{Comparison with Baselines}
Table \ref{comparison_with_baselines} presents a detailed comparison of our proposed MTKD method against several baselines for SER in English, Finnish, and French. 
The performance is reported as the mean of five splits for the IEMOCAP dataset and as the mean of the best and worst splits for the FESC dataset.
A more comprehensive analysis was conducted on the IEMOCAP dataset, given its status as a widely used benchmark, whereas the investigation of the FESC dataset was limited to the best and worst splits to optimize computational resources and time.
Empirical results indicates that while FT and KD yield promising outcomes, MTKD demonstrates superior generalization and robustness, achieving the highest average scores with a substantial amount of data. 
In the monolingual setup, MTKD with a monolingual student (\textit{MTKD-Mono.}) achieves the highest UR and WR in the IEMOCAP and FESC datasets. Likewise, in the multilingual setup, MTKD with a multilingual student (\textit{MTKD-Multi.}) attains the highest UR and WR scores.

\renewcommand{\arraystretch}{1.}
\begin{table}[h!]
\caption{Comparison of our proposed methods with baselines for SER in English, Finnish, and French. 
}
\centering
\resizebox{0.95\columnwidth}{!}{
\begin{tabular}{ccccccc}
\hline
{} & \multicolumn{2}{|c}{IEMOCAP} & \multicolumn{2}{|c}{FESC} & \multicolumn{2}{|c|}{CaFE} \\ 
\cline{2-3}  \cline{4-5}  \cline{6-7}
{} & {UR} & {WR} & {UR} & {WR} & {UR} & {WR} \\ \hline

{FT-Mono.} & {{71.2}} & {{70.1}} & {{59.5}} & {{62.0}} & {78.1} & {78.6} \\

{KD-Mono.} & {72.2} & {71.1} & {62.7} & {67.8} & {\textbf{82.3}} & {\textbf{79.8}} \\

{\textbf{MTKD-Mono.}} & \textbf{72.5} & \textbf{71.2} & \textbf{62.9} & \textbf{68.1} & {78.1} & {75.0} \\

\hline

{FT-Multi.} & {68.2} & {71.0} & {62.7} & {64.3} & \textbf{77.1} & \textbf{79.8} \\

{\textbf{MTKD-Multi.}} & \textbf{71.9} & \textbf{72.9} & \textbf{63.4} & \textbf{66.1} & {73.4} & {72.3} \\

\hline

\end{tabular}
}

\label{comparison_with_baselines}
\end{table}

The \textit{MTKD-Multi.} method demonstrates consistent performance across evaluations. The impact of dataset size on its effectiveness is evident in English SER with the IEMOCAP dataset. Given its relatively large training set (5.7 hours), IEMOCAP exhibits stable performance between the best and average results (Table~\ref{performance_on_iemocap}). Furthermore, while \textit{MTKD-Mono.} achieves a marginal improvement over baseline methods in the monolingual setup, \textit{MTKD-Multi.} outperforms the baseline by a significant margin, highlighting its superior ability to leverage multilingual training for enhanced performance.
However, MTKD-Mono. achieves the highest UR score across compared monolingual and multilingual methods.
In contrast, Finnish SER using the FESC dataset, which contains 3.33 hours of training data, shows a performance drop. For French SER, with only 0.5 hours of training data, \textit{KD-Mono.} demonstrates superior performance compared to other methods. The low performance of \textit{MTKD-Multi.} in French SER is expected, as French data is a minority group in each training batch. As a result, the teacher selected based on the cosine similarity scores is often not the one specialized in French SER, leading to less effective learning for French compared to English and Finnish SER.

\subsubsection{Comparison with Existing State-of-the-Art (SOTA)}
Table \ref{comparison_with_existing_methods} presents a comparative analysis of various SOTA methods for English SER using the IEMOCAP dataset. 
Our proposed monolingual and multilingual MTKD methods outperform existing SOTA approaches in terms of WA and UR, respectively.
However, the slightly higher UA score reported in \cite{zhao2023hierarchical} is anticipated, given that their approach utilizes 20\% of the entire dataset, including duplicated data.

\renewcommand{\arraystretch}{1.}
\begin{table}[h!]
\caption{
Performance comparison of our MTKD method against SOTA methods on the IEMOCAP dataset. Paradigm CKD denotes Cubic Knowledge Distillation. $\textbf{*}$ indicates redundancy.
}
\centering

\resizebox{0.95\columnwidth}{!}{
\begin{tabular}{lllll}
\hline
{{Method}} & {{Paradigm}} & {{WA}} & {{UA}} & {{UR}} \\ \hline

{Wav2vec2-PT~\cite{pepino2021emotion}} & {FT} & {---} & {---} & {67.2}  \\

{DKDFMH~\cite{zhao2023hierarchical}} & {KD} & {---} & {\textbf{77.1$^\textbf{*}$}} & {---}\\

{P-TAPT~\cite{chen2023exploring}} & {FT} & {---} & {74.4} & {---}  \\

{Vesper-4~\cite{chen2024vesper}} & {KD} & {68.4} & {69.3} & {---} \\

{CubicKD~\cite{lou2024cubic}} & {CKD} & {63.3} & {---} & {---} \\

\hline

{\textbf{MTKD-Mono.}} & {MTKD} & {\textbf{76.0}} & {76.2} & \textbf{72.5} \\ 

{\textbf{MTKD-Multi.}} & {MTKD} & {74.8} & {74.9} & {{71.9}} \\ \hline

\end{tabular}
}

\label{comparison_with_existing_methods}
\end{table}

\subsection{Qualitative Analysis}

\begin{figure}[h!]
  \begin{center}
    \includegraphics[width=\columnwidth]{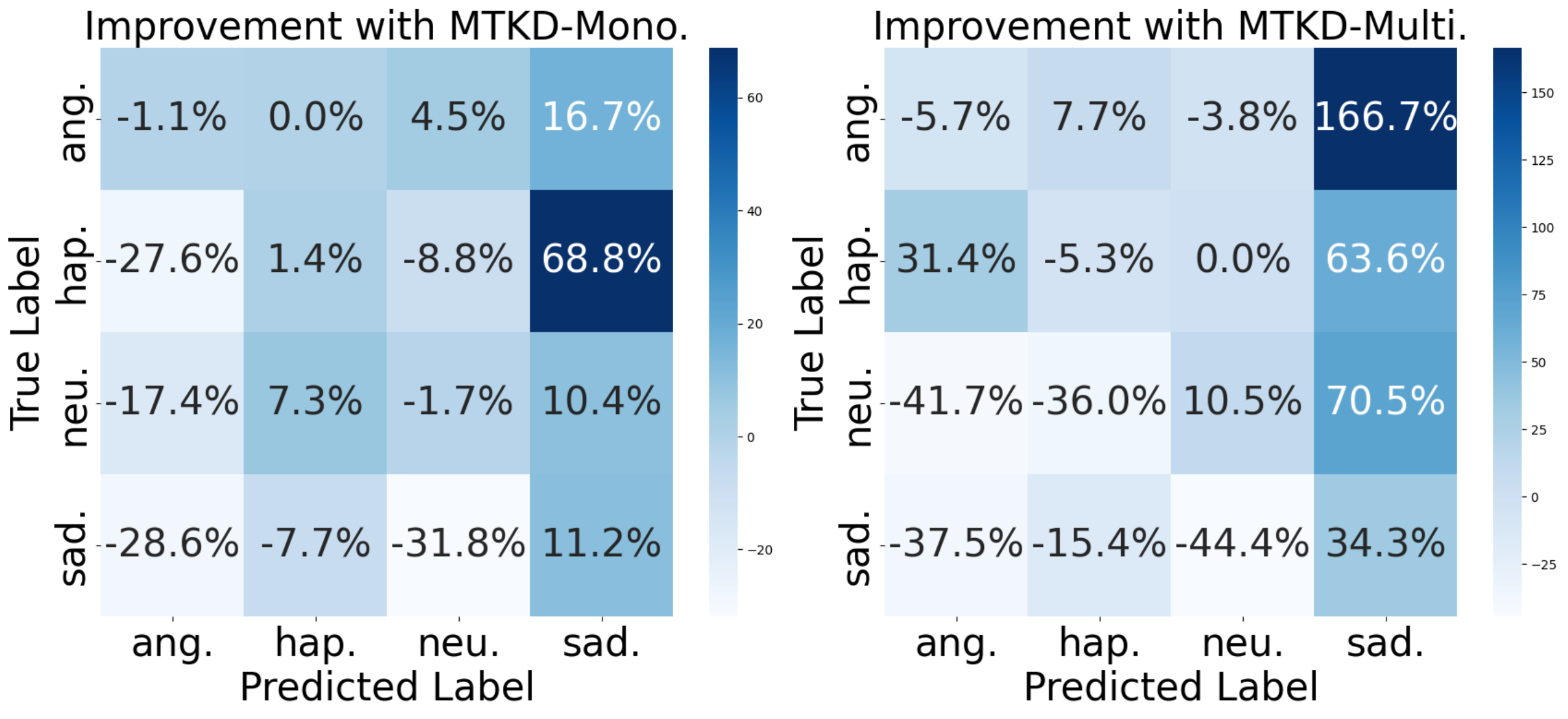} 
  \end{center}
  \caption{
  Performance improvement of MTKD-Mono. over FT-Mono. on monolingual set \textbf{(Left)} and of MTKD-Multi. over FT-Multi. on multilingual set \textbf{(Right)}, respectively.
  } 
  \label{fig:cms_iemocap}
\end{figure}

Figure \ref{fig:cms_iemocap} delineates the average inter-class performance of various SER methods. The comparison includes monolingual (English) and multilingual methods, highlighting the impact of different training approaches. It clarifies that integrating multilingual data enhances overall system effectiveness.

Our proposed method, MTKD-Multi., outperforms other approaches in emotion recognition by reducing misclassification numbers. Compared to the monolingual fine-tuning method (FT-Mono.), which struggles with identifying neutral emotions, and the monolingual knowledge-distillation method (KD-Mono.), which still show confusion between happiness and neutral, our method (MTKD-Mono.) demonstrates superior accuracy and generalization, especially for sadness. The multilingual fine-tuning method (FT-Multi.) improves performance but has issues between neutral and sadness. However, our MTKD-Multi. method enhances overall system effectiveness by improving performance for sadness and neutral emotions.

\subsection{Error Analysis}
The confusion matrices in Figure~\ref{fig:cms_iemocap} reveals that the MTKD-Mono. method has difficulty distinguishing between Happiness and Neutral classes, while the MTKD-Multi. method struggles with Anger class, likely due to being minority classes. Despite these challenges, MTKD-Mono. performs better in reducing misclassifications between Anger and Sadness, and MTKD-Multi. excels in reducing errors between Neutral and Sadness. In terms of recall, MTKD-Mono. improves scores for Happiness and Anger, whereas MTKD-Multi. boosts scores for Neutral and Sadness. Overall, MTKD-Multi. shows superior performance by reducing misclassification rates and improving recall, as highlighted by its higher average UR and WR scores across languages and enhanced performance in English SER. This indicates that leveraging multilingual data and multi-teacher knowledge distillation provides a significant advantage over traditional monolingual methods.

\section{Conclusion}

This study demonstrates the effectiveness of the MTKD approach in markedly enhancing the performance of multilingual SER systems. By leveraging multiple monolingual teacher models, our method facilitates the transfer of cross-linguistic emotional knowledge, outperforming standard FT and conventional KD methods across various languages and datasets. This highlights the importance of optimizing emotional knowledge transfer for improved multilingual emotional understanding.
The quantitative analysis of the IEMOCAP dataset shows that \textit{MTKD-Mono} achieves the highest mean UR of 72.5, while \textit{MTKD-Multi} attains the highest mean WR of 72.9. Furthermore, in the FESC dataset, \textit{MTKD-Mono} achieves the highest WR of 68.1, and \textit{MTKD-Multi} attains the highest UR of 63.4 among all compared methods. 
These results validate that training on out-of-distribution data enhances generalization, even with limited in-domain data.
However, our method currently relies on homogeneous teachers. Moreover, the associated computational demands and challenges in selecting teacher models underscore areas that require further improvement. 
In future studies, we will explore heterogeneous teachers, alternative similarity metrics, and broader linguistic and cultural contexts to develop a more robust and versatile SER system.

 \section{Acknowledgements}
 The computational resources were provided by Aalto ScienceIT. The authors are grateful for the Academy of Finland project funding number 345790 in ICT 2023 programme's project "Understanding speech and scene with ears and eyes" and the Business Finland project LAREINA under Grant 7817/31/2022.

\bibliographystyle{IEEEtran}
\bibliography{mybib}

\begin{thebibliography}{10}
\providecommand{\url}[1]{#1}
\csname url@samestyle\endcsname
\providecommand{\newblock}{\relax}
\providecommand{\bibinfo}[2]{#2}
\providecommand{\BIBentrySTDinterwordspacing}{\spaceskip=0pt\relax}
\providecommand{\BIBentryALTinterwordstretchfactor}{4}
\providecommand{\BIBentryALTinterwordspacing}{\spaceskip=\fontdimen2\font plus
\BIBentryALTinterwordstretchfactor\fontdimen3\font minus \fontdimen4\font\relax}
\providecommand{\BIBforeignlanguage}[2]{{%
\expandafter\ifx\csname l@#1\endcsname\relax
\typeout{** WARNING: IEEEtran.bst: No hyphenation pattern has been}%
\typeout{** loaded for the language `#1'. Using the pattern for}%
\typeout{** the default language instead.}%
\else
\language=\csname l@#1\endcsname
\fi
#2}}
\providecommand{\BIBdecl}{\relax}
\BIBdecl

\bibitem{elsayed2022speech}
N.~Elsayed, Z.~ElSayed, N.~Asadizanjani, M.~Ozer, A.~Abdelgawad, and M.~Bayoumi, ``Speech emotion recognition using supervised deep recurrent system for mental health monitoring,'' in \emph{2022 IEEE 8th World Forum on Internet of Things (WF-IoT)}.\hskip 1em plus 0.5em minus 0.4em\relax IEEE, 2022, pp. 1--6.

\bibitem{grosz2023discovering}
T.~Gr{\'o}sz, A.~Virkkunen, D.~Porjazovski, and M.~Kurimo, ``Discovering relevant sub-spaces of bert, wav2vec 2.0, electra and vit embeddings for humor and mimicked emotion recognition with integrated gradients,'' in \emph{Proceedings of the 4th on Multimodal Sentiment Analysis Challenge and Workshop: Mimicked Emotions, Humour and Personalisation}, 2023, pp. 27--34.

\bibitem{zhang2022confidence}
H.~Zhang, D.~Chen, and C.~Wang, ``Confidence-aware multi-teacher knowledge distillation,'' in \emph{ICASSP 2022-2022 IEEE International Conference on Acoustics, Speech and Signal Processing (ICASSP)}.\hskip 1em plus 0.5em minus 0.4em\relax IEEE, 2022, pp. 4498--4502.

\bibitem{zhang2023adaptive}
------, ``Adaptive multi-teacher knowledge distillation with meta-learning,'' in \emph{2023 IEEE International Conference on Multimedia and Expo (ICME)}.\hskip 1em plus 0.5em minus 0.4em\relax IEEE, 2023, pp. 1943--1948.

\bibitem{anand2023multi}
S.~Anand, N.~K. Devulapally, S.~D. Bhattacharjee, and J.~Yuan, ``Multi-label emotion analysis in conversation via multimodal knowledge distillation,'' in \emph{Proceedings of the 31st ACM International Conference on Multimedia}, 2023, pp. 6090--6100.

\bibitem{baevski2020wav2vec}
A.~Baevski, Y.~Zhou, A.~Mohamed, and M.~Auli, ``wav2vec 2.0: A framework for self-supervised learning of speech representations,'' \emph{Advances in neural information processing systems}, vol.~33, pp. 12\,449--12\,460, 2020.

\bibitem{hsu2021hubert}
W.-N. Hsu, B.~Bolte, Y.-H.~H. Tsai, K.~Lakhotia, R.~Salakhutdinov, and A.~Mohamed, ``Hubert: Self-supervised speech representation learning by masked prediction of hidden units,'' \emph{IEEE/ACM transactions on audio, speech, and language processing}, vol.~29, pp. 3451--3460, 2021.

\bibitem{chen2022wavlm}
S.~Chen, C.~Wang, Z.~Chen, Y.~Wu, S.~Liu, Z.~Chen, J.~Li, N.~Kanda, T.~Yoshioka, X.~Xiao \emph{et~al.}, ``Wavlm: Large-scale self-supervised pre-training for full stack speech processing,'' \emph{IEEE Journal of Selected Topics in Signal Processing}, vol.~16, no.~6, pp. 1505--1518, 2022.

\bibitem{pepino2021emotion}
L.~Pepino, P.~Riera, and L.~Ferrer, ``Emotion recognition from speech using wav2vec 2.0 embeddings,'' \emph{arXiv preprint arXiv:2104.03502}, 2021.

\bibitem{grosz2022wav2vec2}
T.~Gr{\'o}sz, D.~Porjazovski, Y.~Getman, S.~Kadiri, and M.~Kurimo, ``Wav2vec2-based paralinguistic systems to recognise vocalised emotions and stuttering,'' in \emph{Proceedings of the 30th ACM International Conference on Multimedia}, 2022, pp. 7026--7029.

\bibitem{chen2023exploring}
L.-W. Chen and A.~Rudnicky, ``Exploring wav2vec 2.0 fine tuning for improved speech emotion recognition,'' in \emph{ICASSP 2023-2023 IEEE International Conference on Acoustics, Speech and Signal Processing (ICASSP)}.\hskip 1em plus 0.5em minus 0.4em\relax IEEE, 2023, pp. 1--5.

\bibitem{sharma2022multi}
M.~Sharma, ``Multi-lingual multi-task speech emotion recognition using wav2vec 2.0,'' in \emph{ICASSP 2022-2022 IEEE International Conference on Acoustics, Speech and Signal Processing (ICASSP)}.\hskip 1em plus 0.5em minus 0.4em\relax IEEE, 2022, pp. 6907--6911.

\bibitem{chakhtouna2024unveiling}
A.~Chakhtouna, S.~SEKKATE, and A.~Abdellah, ``Unveiling embedded features in wav2vec2 and hubert msodels for speech emotion recognition,'' \emph{Procedia Computer Science}, vol. 232, pp. 2560--2569, 2024.

\bibitem{gao2022multi}
L.~Gao, K.~Xu, H.~Wang, and Y.~Peng, ``Multi-representation knowledge distillation for audio classification,'' \emph{Multimedia Tools and Applications}, vol.~81, no.~4, pp. 5089--5112, 2022.

\bibitem{hao2024one}
Z.~Hao, J.~Guo, K.~Han, Y.~Tang, H.~Hu, Y.~Wang, and C.~Xu, ``One-for-all: Bridge the gap between heterogeneous architectures in knowledge distillation,'' \emph{Advances in Neural Information Processing Systems}, vol.~36, 2024.

\bibitem{zhao2023hierarchical}
Z.~Zhao, H.~Wang, H.~Wang, and B.~Schuller, ``Hierarchical network with decoupled knowledge distillation for speech emotion recognition,'' in \emph{ICASSP 2023-2023 IEEE International Conference on Acoustics, Speech and Signal Processing (ICASSP)}.\hskip 1em plus 0.5em minus 0.4em\relax IEEE, 2023, pp. 1--5.

\bibitem{yun2023end}
H.-I. Yun and J.-S. Park, ``End-to-end emotional speech recognition using acoustic model adaptation based on knowledge distillation,'' \emph{Multimedia Tools and Applications}, vol.~82, no.~15, pp. 22\,759--22\,776, 2023.

\bibitem{sridhar2020ensemble}
K.~Sridhar and C.~Busso, ``Ensemble of students taught by probabilistic teachers to improve speech emotion recognition.'' in \emph{INTERSPEECH}, 2020, pp. 516--520.

\bibitem{takashima2018investigation}
R.~Takashima, S.~Li, and H.~Kawai, ``An investigation of a knowledge distillation method for ctc acoustic models,'' in \emph{2018 IEEE International Conference on Acoustics, Speech and Signal Processing (ICASSP)}.\hskip 1em plus 0.5em minus 0.4em\relax IEEE, 2018, pp. 5809--5813.

\bibitem{fukuda2017efficient}
T.~Fukuda, M.~Suzuki, G.~Kurata, S.~Thomas, J.~Cui, and B.~Ramabhadran, ``Efficient knowledge distillation from an ensemble of teachers.'' in \emph{Interspeech}, 2017, pp. 3697--3701.

\bibitem{yang2020model}
Z.~Yang, L.~Shou, M.~Gong, W.~Lin, and D.~Jiang, ``Model compression with two-stage multi-teacher knowledge distillation for web question answering system,'' in \emph{Proceedings of the 13th International Conference on Web Search and Data Mining}, 2020, pp. 690--698.

\bibitem{ren2023fast}
Z.~Ren, T.~T. Nguyen, Y.~Chang, and B.~W. Schuller, ``Fast yet effective speech emotion recognition with self-distillation,'' in \emph{ICASSP 2023-2023 IEEE International Conference on Acoustics, Speech and Signal Processing (ICASSP)}.\hskip 1em plus 0.5em minus 0.4em\relax IEEE, 2023, pp. 1--5.

\bibitem{huang2023ensemble}
K.-P. Huang, T.-H. Feng, Y.-K. Fu, T.-Y. Hsu, P.-C. Yen, W.-C. Tseng, K.-W. Chang, and H.-Y. Lee, ``Ensemble knowledge distillation of self-supervised speech models,'' in \emph{ICASSP 2023-2023 IEEE International Conference on Acoustics, Speech and Signal Processing (ICASSP)}.\hskip 1em plus 0.5em minus 0.4em\relax IEEE, 2023, pp. 1--5.

\bibitem{chen2024vesper}
W.~Chen, X.~Xing, P.~Chen, and X.~Xu, ``Vesper: A compact and effective pretrained model for speech emotion recognition,'' \emph{IEEE Transactions on Affective Computing}, 2024.

\bibitem{busso2008iemocap}
C.~Busso, M.~Bulut, C.-C. Lee, A.~Kazemzadeh, E.~Mower, S.~Kim, J.~N. Chang, S.~Lee, and S.~S. Narayanan, ``Iemocap: Interactive emotional dyadic motion capture database,'' \emph{Language resources and evaluation}, vol.~42, pp. 335--359, 2008.

\bibitem{airas2006emotions}
M.~Airas and P.~Alku, ``Emotions in vowel segments of continuous speech: analysis of the glottal flow using the normalised amplitude quotient,'' \emph{Phonetica}, vol.~63, no.~1, pp. 26--46, 2006.

\bibitem{gournay2018canadian}
P.~Gournay, O.~Lahaie, and R.~Lefebvre, ``A canadian french emotional speech dataset,'' in \emph{Proceedings of the 9th ACM multimedia systems conference}, 2018, pp. 399--402.

\bibitem{lou2024cubic}
Z.~Lou, S.~Otake, Z.~Li, R.~Kawakami, and N.~Inoue, ``Cubic knowledge distillation for speech emotion recognition,'' in \emph{ICASSP 2024-2024 IEEE International Conference on Acoustics, Speech and Signal Processing (ICASSP)}.\hskip 1em plus 0.5em minus 0.4em\relax IEEE, 2024, pp. 5705--5709.

\end{thebibliography}

\end{document}